\documentclass{article}
\usepackage{spconf,amsmath,graphicx}
\usepackage{booktabs}
\usepackage{pifont}
\usepackage{amssymb}
\usepackage{algorithmic}
\usepackage{graphicx}
\usepackage{textcomp}
\usepackage{booktabs}
\usepackage{colortbl}
\usepackage[table,xcdraw]{xcolor}
\usepackage{multirow}
\usepackage{marvosym}
\usepackage{enumitem}
\setlist{nosep, leftmargin=14pt}

\usepackage{mwe} 

\title{MARIO: A Mixed Annotation Framework for Polyp Segmentation}

\name{Haoyang Li$^{1\ast}$\thanks{Equal contributions.} 
\qquad Yiwen Hu$^{3\ast}$ 
\qquad Jun Wei$^{4\ast}$ 
\qquad Zhen Li$^{1,2}$\sthanks{Corresponding Author: Prof. Zhen Li, Email: lizhen@cuhk.edu.cn}}

\address{$^1$  FNii, CUHK-Shenzhen \\
 $^2$ SSE, CUHK-Shenzhen \\
 $^3$ South China Hospital, Shenzhen University  \\
 $^4$ CSSE, Shenzhen University      
 }

\begin{document}
%
\maketitle
\begin{abstract}
Existing polyp segmentation models are limited by high labeling costs and the small size of datasets. Additionally, vast polyp datasets remain underutilized because these models typically rely on a single type of annotation. To address this dilemma, we introduce \textbf{MARIO}, a mixed supervision model designed to accommodate various annotation types, significantly expanding the range of usable data. MARIO learns from underutilized datasets by incorporating five forms of supervision: pixel-level, box-level, polygon-level, scribble-level, and point-level. Each form of supervision is associated with a tailored loss that effectively leverages the supervision labels while minimizing the noise. This allows MARIO to move beyond the constraints of relying on a single annotation type. Furthermore, MARIO primarily utilizes dataset with weak and cheap annotations, reducing the dependence on large-scale, fully annotated ones. Experimental results across five benchmark datasets demonstrate that MARIO consistently outperforms existing methods, highlighting its efficacy in balancing trade-offs between different forms of supervision and maximizing polyp segmentation performance.
\end{abstract}

\section{Introduction}
\label{sec:intro}
Colorectal cancer remains a significant global health challenge, with early detection and treatment of polyps critical for improving patient outcomes. Recent advances in deep learning have significantly improved polyp segmentation, with models such as U-Net and its variants~\cite{unet,pranet,sanet} demonstrating strong performance. Transformer-based architectures, such as Polyp-Pvt~\cite{polyppvt}, have further enhanced segmentation accuracy. However, these methods are constrained by the high cost of annotation and the limited scale of existing datasets.

To address these challenges, weakly supervised methods like WeakPolyp~\cite{wei2023weakpolyp} leverage low-cost bounding box annotations to reduce labeling efforts but fail to exploit the diversity of available annotation types. To overcome this limitation, we propose MARIO, a mixed-supervision framework that integrates five annotation types—pixel-level, polygon-level, bounding box-level, scribble-level, and point-level. This unified architecture enhances dataset utility, reduces labeling burdens, and provides a scalable, practical solution for clinical applications.

Despite its advanced design, weak labels introduce noise that can hinder model performance. MARIO addresses this challenge with specialized loss functions tailored to the characteristics of each annotation type. For example, pixel-level annotations, offering fine-grained detail, benefit from metrics like binary cross-entropy and Dice loss, while bounding box annotations rely on mask-to-box transformations and consistency losses to align predictions with spatial context and address label inaccuracies. These tailored losses enhance training robustness, enabling effective learning from both detailed and coarse labels. This diverse annotation strategy optimizes data usage, addressing data scarcity while mitigating limitations associated with specific annotation types. By seamlessly incorporating multiple annotation formats, MARIO expands usable training data, improves generalization across datasets, and ensures practical adoption by accommodating varied annotation preferences of medical professionals.

In conclusion, MARIO provides a scalable solution for colorectal polyp segmentation by unifying five annotation types within a single framework. Tailored loss functions mitigate noise and enhance training robustness, enabling effective learning from diverse annotations. This flexible approach optimizes data usage by leveraging diverse annotations, enhances generalizability across datasets through robust training, and advances colorectal cancer screening by improving segmentation accuracy and diagnostic precision, ensuring practical applicability in clinical workflows.

\begin{figure*}[t]
    \centering
    \includegraphics[width=\linewidth]{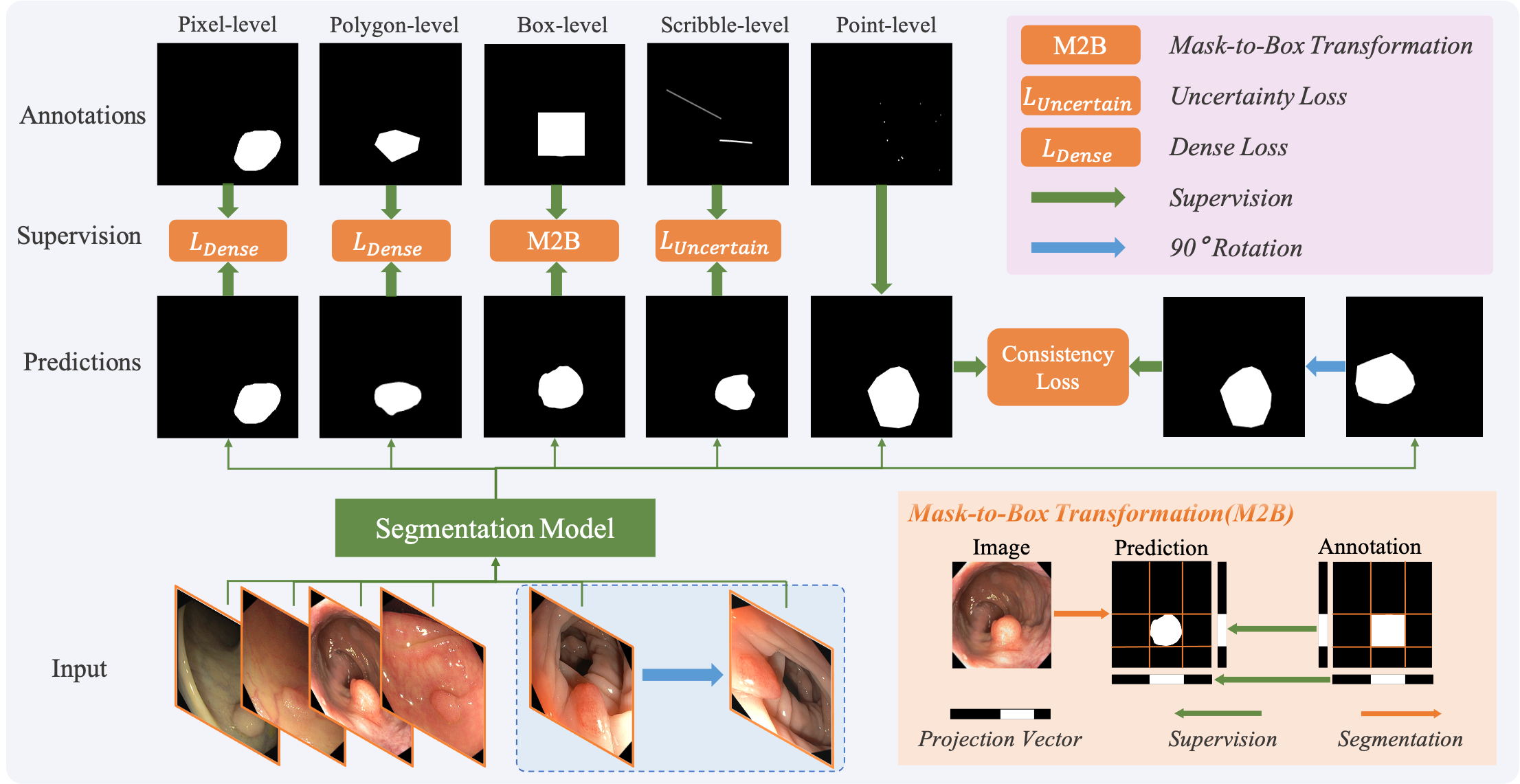}
    \caption{\textbf{Illustration of our MARIO framework.} 
    }
    \label{fig:pipeline}
\end{figure*}

\section{Methodology}
\label{sec:Methodology}
\subsection{Transformer-based Polyp Segmentation Model}
\begin{table*}[t]
    \centering
    \caption{Performance comparison with different polyp segmentation models. The \textcolor{red}{red} column represents the weighted average (wAVG) performance of different testing datasets. Next to the dataset name is the image quantity of each dataset.}
    \label{tab:performance}
    \renewcommand\tabcolsep{4.6pt}
    \begin{tabular}{lcccccccccccc}
    \toprule
        & \multicolumn{2}{c}{\textbf{ColonDB} (380)} & \multicolumn{2}{c}{\textbf{Kvasir} (100)} & \multicolumn{2}{c}{\textbf{ClinicDB} (62)} & \multicolumn{2}{c}{\textbf{EndoScene} (60)} & \multicolumn{2}{c}{\textbf{ETIS} (196)} & \multicolumn{2}{c}{\textcolor{red}{\textbf{wAVG} (798)}} \\
        \cmidrule(l){2-3}\cmidrule(l){4-5}\cmidrule(l){6-7}\cmidrule(l){8-9}\cmidrule(l){10-11}\cmidrule(l){12-13}
    \multirow{-3}{*}{\textbf{Methods}}   & Dice & IoU  & Dice & IoU  & Dice & IoU  & Dice & IoU  & Dice & IoU  & \textcolor{red}{Dice} & \textcolor{red}{IoU} \\ 
    \hline
    U-Net~\cite{unet}             & 51.2\% & 44.4\% & 81.8\% & 74.6\% & 82.3\% & 75.0\% & 71.0\% & 62.7\% & 39.8\% & 33.5\% & \textcolor{red}{56.1\%} & \textcolor{red}{49.3\%}\\
    PraNet~\cite{pranet}          & 70.9\% & 64.0\% & 89.8\% & 84.0\% & 89.9\% & 84.9\% & 87.1\% & 79.7\% & 62.8\% & 56.7\% & \textcolor{red}{74.0\%} & \textcolor{red}{67.5\%}\\
    SANet~\cite{sanet}            & 75.3\% & 67.0\% & 90.4\% & 84.7\% & 91.6\% & 85.9\% & 88.8\% & 81.5\% & 75.0\% & 65.4\% & \textcolor{red}{79.4\%} & \textcolor{red}{71.4\%}\\  
    Polyp-Pvt~\cite{polyppvt}     & 80.8\% & 72.7\% & 91.7\% & 86.4\% & 93.7\% & 88.9\% & 90.0\% & 83.3\% & 78.7\% & 70.6\% & \textcolor{red}{83.3\%} & \textcolor{red}{76.0\%}\\
    LDNet~\cite{ldnet}            & 79.4\% & 71.5\% & 91.2\% & 85.5\% & 92.3\% & 87.2\% & 89.3\% & 82.6\% & 77.8\% & 70.7\% & \textcolor{red}{82.2\%} & \textcolor{red}{75.1\%}\\
    HSNet~\cite{zhang2022hsnet}   & 81.0\% & 73.5\% & 92.6\% & 87.7\% & 94.8\% & 90.5\% & 90.3\% & 83.9\% & 80.8\% & 73.4\% & \textcolor{red}{84.2\%} & \textcolor{red}{77.4\%}\\ 
    UCFA-Net~\cite{UCFANet}   & 82.3\% & 74.1\% & 91.7\% & 86.8\% & 93.4\% & 87.0\% & 89.7\% & 83.0\% & 82.3\% & 74.3\% & \textcolor{red}{84.8\%} & \textcolor{red}{77.3\%}\\ 
    CAFE-Net~\cite{LIU2024121754}   & 82.0\% & 74.0\% & 93.3\% & 88.9\% & 94.3\% & 89.9\% & 90.1\% & 83.4\% & 82.2\% & 73.8\% & \textcolor{red}{84.9\%} & \textcolor{red}{77.7\%}\\ 
    \rowcolor{black!10}
    \textbf{MARIO (Ours)}         & 82.8\% & 74.5\% & 91.7\% & 86.2\% & 91.9\% & 86.7\% & 90.7\% & 83.9\% & 85.1\% & 77.6\% & \textbf{\textcolor{red}{85.8\%}} & \textbf{\textcolor{red}{78.3\%}}\\
    \bottomrule
\end{tabular}
\end{table*}
MARIO is a transformer-based weakly supervised segmentation model that uses PVTv2-B2 \cite{wang2021pvtv2} as its backbone. It initially processes input images with dimensions height ${H}$ and width ${W}$, extracting feature maps at four distinct scales as sizes of $\frac{1}{2}$, $\frac{1}{4}$, $\frac{1}{8}$, $\frac{1}{16}$ of the input images. To reduce the computational cost, we focus on the three lower-resolution feature maps, then apply a $1 \times 1$ convolutional layer to ensure consistency in channel dimensions across these feature maps. Before the prediction stage, bilinear upsampling is employed to align the spatial dimensions of the feature maps, followed by a final $1 \times 1$ convolutional layer for prediction. 

To maximize the range of usable data, MARIO incorporates five types of annotations: pixel-level \((M_{pixel})\), polygon-level \((M_{polygon})\), box-level \((M_{box})\), scribble-level \((M_{scribble})\), and point-level \((M_{points})\). Fig.~\ref{fig:pipeline} illustrates the pipeline of the proposed method, where each type of data is input into the segmentation model to generate corresponding predictions. This approach facilitates simultaneous learning from multiple annotation types within a single framework, simplifying the training process and enabling the model to leverage the complementary strengths of various annotations, ultimately resulting in superior segmentation performance.

\subsection{Loss Function Definition}
In MARIO, the supervision losses are carefully designed to align with specific types of annotations, which can be categorized into four groups: dense supervision, box supervision, scribble supervision, and point supervision.

\textbf{Dense Supervision.} 
Dense supervision, building on prior studies \cite{pranet, sanet}, has proven effective in supervised learning tasks by leveraging a combination of binary cross-entropy\((\mathcal{L}_{BCE})\)  and Dice loss \((\mathcal{L}_{Dice})\). Dice loss maximizes spatial overlap, effectively addressing class imbalance, while BCE ensures pixel-level precision, particularly at boundaries. Together, these losses complement each other to provide robust training for accurate and detailed segmentation, making them particularly suitable for pixel-level and polygon-level annotations. In dense supervision, the loss is denoted as \(\mathcal{L}_{\text{pixel}}\) for pixel-level data and \(\mathcal{L}_{\text{polygon}}\) for polygon-level data.

\textbf{Box Supervision.} For box supervision, we utilize the mask-to-box \((M2B)\) transformation to align the predicted results and labels in the same space, eliminating potential biases from box shapes. Specifically, \(M2B\) consists of two main procedures: a projection step that projects the predicted mask into two vectors along the row and column directions, and a back-projection step that reconstructs a box-shaped mask from these vectors. This projection mechanism eliminates inconsistencies in shapes between predictions and labels, allowing supervision to focus more on the target's location rather than its shape, thereby avoiding misleading noise. The supervision loss is denoted as \(\mathcal{L}_{\text{box}}\). 

\textbf{Scribble Supervision.} Scribble supervision is sparse, leaving most areas of the image unannotated and without effective supervision, which increases uncertainty in the predictions. To address this challenge, we propose the uncertainty loss \(\mathcal{L}_{Uncertain}\), or \(\mathcal{L}_{\text{scribble}}\) to mitigate the uncertainty introduced by scribble annotations. \(\mathcal{L}_{Uncertain}\) specifically tackles the lack of ground truth for the majority of pixels, preventing the model from becoming overconfident and making erroneous predictions in unlabeled regions. By combining CE loss for labeled pixels with uncertainty loss for unlabeled ones, the model effectively utilizes sparse annotations while maintaining flexibility and avoiding bias in the unlabeled areas, making it ideal supervision for such a weakly supervised scenario. The uncertainty loss is defined in Eq.~\ref{equ:uncertain}, where \(P_s^i\) represents the predicted probability for the sample.
\begin{equation}
    \mathcal{L}_{\text{scribble}}=\mathcal{L}_{\text{Uncertain}} = \min\left(-\log(P_s^i),-\log(1 - P_s^i) \right).
    \label{equ:uncertain}
\end{equation}
\textbf{Point Supervision.} Point annotations are the sparsest form of labeling, introducing significant noise into the model. To mitigate this impact, we propose a consistency loss \(\mathcal{L}_{\text{points}}\) that constrains the consistency of output results. Specifically, we perform a 90-degree rotation on each input image, then reverse the rotation to compare the consistency between the two predictions. This consistency loss is calculated by computing the MSE loss between the model’s predictions on the original images and those on their 90-degree rotated counterparts. By enforcing consistency between these predictions, our method effectively reduces the influence of noise from sparse point annotations, enhancing the overall robustness and accuracy of the segmentation model. The total loss is derived as the cumulative sum of individual supervision losses, each contributing to the overall optimization process:
\begin{equation}
    \mathcal{L}_{\text{total}} = \mathcal{L}_{\text{pixel}} + \mathcal{L}_{\text{box}} + \mathcal{L}_{\text{polygon}} + \mathcal{L}_{\text{scribble}} + \mathcal{L}_{\text{points}}.
    \label{loss:total_alt}
\end{equation}


\section{Experiment}
\label{sec:Experiment}
\subsection{Dataset and Implementation Detail}
This study leverages eight datasets: Kvasir, CVC-ClinicDB, CVC-ColonDB, EndoScene, ETIS, SUN-SEG, LDPolypVideo, and PolypGen. A total of 1,451 pixel-level annotated images from the first five datasets were used for training, with the remainder for testing. LDPolypVideo provided 33,884 box-annotated samples, while SUN-SEG's 49,136 samples were evenly split for polygon and scribble annotations. PolypGen contributed 1,412 images with point annotations (5 foreground and 5 background points). This diverse annotation framework strengthens model training and generalization.

In this study, we employed a neural network architecture based on the Pyramid Vision Transformer v2 (PVTv2) backbone \cite{wang2021pvtv2} for the MARIO framework. The implementation was conducted using PyTorch 2.3.0. Input images were resized to random dimensions selected from a predefined set using bilinear interpolation. The model optimization was performed using SGD with a momentum of 0.9 and an initial learning rate of 0.05. We configured the batch size to 4, and the training process consisted of 60,000 iteration steps. Dice coefficient and Intersection over Union (IoU) are utilized to assess the performance of MARIO, providing a comprehensive evaluation of the model's effectiveness across varying annotation types and conditions

\begin{figure}[t]
    \centering
    \includegraphics[width=\linewidth]{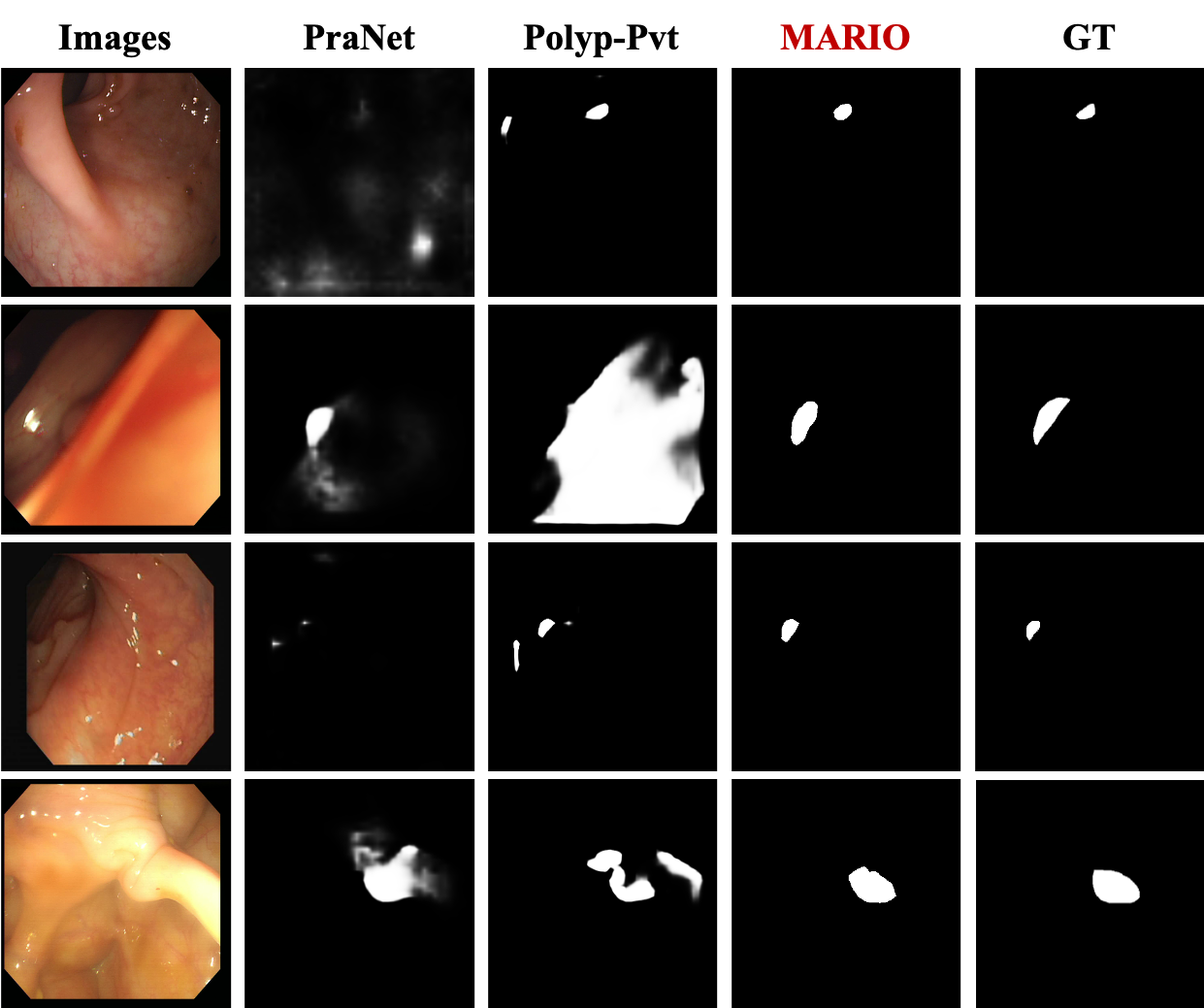}
    \caption{\textbf{Visualization results of our MARIO and other comparison methods. "GT" denotes the ground truth.} 
    }
    \label{fig:vis}
\end{figure}

\newcommand{\cmark}{\checkmark}
\newcommand{\xmark}{}
\begin{table}[t]
    \centering
    \caption{The ablation study for different loss}
    \label{tab:performance_metrics}
    \begin{tabular}{ccccc}
        \toprule
        \textbf{BCE} & \textbf{Uncertain} & \textbf{Consistency} & \textbf{Dice (\%)} & \textbf{IoU (\%)} \\ 
        \midrule
        \cmark & \xmark & \xmark & 85.24          & 77.91 \\
        \cmark & \cmark & \xmark & 85.37          & 78.23 \\
        \cmark & \cmark & \cmark & \textbf{85.81} & \textbf{78.39} \\
        \bottomrule
    \end{tabular}
\end{table}

\subsection{Performance Comparison}
We evaluated our mixed-supervised method MARIO against eight state-of-the-art fully supervised polyp segmentation models across five diverse datasets. As shown in Table~\ref{tab:performance}, MARIO achieved the highest weighted average (wAVG) performance, with a Dice score of 85.8\% and an IoU of 78.3\%, surpassing the second-best model, CAFA-Net, by 1.0\% in Dice and 0.6\% in IoU. MARIO also excelled on individual datasets, achieving 82.8\% Dice on ColonDB, 91.7\% on Kvasir, and 85.1\% on ETIS. These results underscore MARIO's superior segmentation accuracy and robustness, enabled by its mixed supervision approach, which integrates multiple annotation types and outperforms fully supervised methods reliant on costly pixel-level annotations.

\subsection{Ablation Study}
The ablation study presented in Table 2 demonstrates the incremental benefits of incorporating multiple loss functions. Starting with $L_{BCE}$ alone, the model achieves a Dice score of 85.24\% and an IoU of 77.91\%. Introducing the Uncertain loss slightly enhances performance to a Dice score of 85.37\% and an IoU of 78.23\%. Further adding the Consistency loss results in the highest observed metrics, with a Dice score of 85.81\% and an IoU of 78.39\%. These findings indicate that each additional loss component contributes to improved segmentation accuracy, highlighting the effectiveness of a multi-faceted loss strategy in enhancing model performance. 

\section{Conclusion}
\label{sec:Conclusion}

To address the challenge of data scarcity due to high labeling costs, we present MARIO, a mixed-supervised model for polyp segmentation. MARIO unifies five annotation types: pixel-level, polygon-level, box-level, scribble-level, and point-level. This integration maximizes the utilization of existing annotated data, caters to medical professionals' annotation preferences, and improves usability for clinical applications. We also design specific loss functions for each annotation type. These functions adopt the mixing nature of annotations and reduce noise, making the model more robust and practical. Experiments on five diverse datasets show that MARIO outperforms existing methods in polyp segmentation. It effectively optimizes data usage, boosts segmentation accuracy, and meets clinical needs.

\section{COMPLIANCE WITH ETHICAL STANDARDS}
\label{sec:print}
This is a retrospective analysis relying exclusively on publicly available, fully anonymized datasets, and the need for ethical approval was waived.



\bibliographystyle{IEEEbib}
\bibliography{bibtex}

\end{document}